\documentclass[conference]{IEEEtran}
\IEEEoverridecommandlockouts
\usepackage{cite}
\usepackage{amsmath,amssymb,amsfonts}
\usepackage{algorithmic}
\usepackage{graphicx}
\usepackage{textcomp}
\usepackage{xcolor}
\usepackage{url}
\usepackage{xurl}
\def\BibTeX{{\rm B\kern-.05em{\sc i\kern-.025em b}\kern-.08em
    T\kern-.1667em\lower.7ex\hbox{E}\kern-.125emX}}

\usepackage{textcomp} 
\usepackage{amssymb} 
\usepackage{bm} 
\usepackage{booktabs} 
\usepackage{dcolumn} 
\usepackage{multirow} 
\usepackage{graphicx} 
\usepackage[printonlyused]{acronym} 
 \usepackage{tabularx}
\usepackage{array}
\usepackage{rotating}
\usepackage{amssymb}
\usepackage{float}

\usepackage{graphicx}
\usepackage{hyperref}
\usepackage{pdflscape}
\usepackage{afterpage}
\usepackage{longtable}
\usepackage{colortbl}
\usepackage{lipsum}
\usepackage{changepage}

\hypersetup{
 colorlinks,
 linkcolor=blue,
 filecolor=magenta, 
 urlcolor=blue,
 citecolor=blue}
\newcolumntype{C}[1]{>{\centering\arraybackslash}p{#1}}
\begin{document}
\title{Big Data and Deep Learning in Smart Cities: A Comprehensive Dataset for AI-Driven Traffic Accident Detection and Computer Vision Systems
}


\author{\IEEEauthorblockN{Victor Adewopo}
\IEEEauthorblockA{\textit{School of IT} \\
\textit{University of Cincinnati}\\
Cincinnati, USA \\
Adewopva@mail.uc.edu}
\and
\IEEEauthorblockN{Nelly Elsayed}
\IEEEauthorblockA{\textit{School of IT} \\
\textit{University of Cincinnati}\\
Cincinnati, USA \\
elsayeny@ucmail.uc.edu}
\and
\IEEEauthorblockN{Zag Elsayed}
\IEEEauthorblockA{\textit{School of IT} \\
\textit{University of Cincinnati}\\
Cincinnati, USA \\
elsayezs@ucmail.uc.edu}
\and
\IEEEauthorblockN{Murat Ozer}
\IEEEauthorblockA{\textit{School of IT} \\
\textit{University of Cincinnati}\\
Cincinnati, USA \\
ozermm@ucmail.uc.edu}
\and
\IEEEauthorblockN{Constantinos L. Zekios}
\IEEEauthorblockA{\textit{Dept. of Elec. \& Comp. Eng.} \\
\textit{Florida International University}\\
Florida, United States \\
kzekios@fiu.edu}
\and
\IEEEauthorblockN{Ahmed Abdelgawad}
\IEEEauthorblockA{\textit{School of Eng. \& Tech.} \\
\textit{Central Michigan University}\\
Michigan, USA \\
abdel1a@cmich.edu}

\and
\IEEEauthorblockN{Magdy Bayoumi}
\IEEEauthorblockA{\textit{University of Louisiana} \\
\textit{Dept. of Elec.\& Comp. Eng.}\\
Lafayette, Louisiana \\
magdy.bayoumi@louisiana.edu}
}

\maketitle

\begin{abstract}
In the dynamic urban landscape, where interplay of vehicles and pedestrians defines the rhythm of life, integrating advanced technology for safety and efficiency is increasingly crucial. This study delves into the application of cutting-edge technological methods in smart cities, focusing on enhancing public safety through improved traffic accident detection. Action recognition plays a pivotal role in interpreting visual data and tracking object motion such as human pose estimation in video sequences. The challenges of action recognition include variability in rapid actions, limited dataset, and environmental factors such as (Weather, Illumination, and Occlusions). In this paper, we present a novel comprehensive \href{http://ieee-dataport.org/documents/comprehensive-dataset-ai-driven-traffic-accident-detection-and-computer-vision-systems}{dataset} for traffic accident detection. This dataset is specifically designed to bolster computer vision and action recognition systems in predicting and detecting road traffic accidents. We integrated datasets from  wide variety of data sources, road networks, weather conditions, and regions across the globe. This approach is underpinned by empirical studies, aiming to contribute to the discourse on how technology can enhance the quality of life in densely populated areas. 
This research aims to bridge existing research gaps by introducing benchmark \href{http://ieee-dataport.org/documents/comprehensive-dataset-ai-driven-traffic-accident-detection-and-computer-vision-systems}{datasets} that leverage state-of-the-art algorithms tailored for traffic accident detection in smart cities. These \href{http://ieee-dataport.org/documents/comprehensive-dataset-ai-driven-traffic-accident-detection-and-computer-vision-systems}{dataset} is expected to advance academic research and also enhance real-time accident detection applications, contributing significantly to the evolution of smart urban environments. Our study marks a pivotal step towards safer, more efficient smart cities, harnessing the power of AI and machine learning to transform urban living.
\end{abstract}


\section{Introduction}
In urban landscape settings, where the rhythm of life activities is punctuated by the ceaseless motion of vehicles and pedestrians, the role of technology in enhancing safety and efficiency becomes paramount. Our study ventures into the realm of applying advanced technological methods to safeguard individuals in smart cities, a concept that has gained traction in recent years \cite{neirotti2014current}.
The field of computer vision, particularly action recognition, stands at the forefront of this endeavor. Action recognition, a nuanced area within computer vision, is dedicated to the automated identification and interpretation of human activities in video data. The significance spans across various sectors, including security, healthcare, entertainment, and urban management \cite{poppe2010survey}. This discipline involves sophisticated techniques such as detecting and tracking human body parts, pose estimation, and deciphering the actions captured in video sequences \cite{Soomro2014, zhang2019comprehensive}.
Though action recognition is a specific task within the larger framework of computer vision, it shares numerous methodologies with its parent field, such as feature extraction, object detection, and motion tracking. However, it is also confronted with unique challenges like variability in a human pose, changes in illumination, occlusions, and the complexity of actions, which require tailored techniques and models \cite{ahad2012motion, zhang2019comprehensive}.
In the context of smart city development, the study by Neirotti et al.~\cite{neirotti2014current}, analyzing 70 international cities, illuminates the pivotal role of transportation systems. Their research underscores the need for optimized urban logistics; considering traffic dynamics and energy efficiency, this approach enhances urban mobility and steers cities toward sustainable development. The advancement of such intelligent transport systems is a crucial step in evolving smart, sustainable cities, reflecting a harmonious blend of technological innovation and environmental responsibility \cite{neirotti2014current}.
Central to this research is the development of a robust dataset \cite{adewopodataset}, tailored for enhancing computer vision and action recognition systems in the context of traffic accident detection. This paper proposes a novel approach that integrates low-cost intelligent technologies with open-source data to augment safety in smart urban settings. This approach is grounded in empirical studies and aims to contribute to the growing body of knowledge on how technology can be effectively employed to improve the quality of life in densely populated areas \cite{caragliu2013smart}.
In the modern era of technology, there is an ever-growing need for new tools that allow users to manage and utilize information overload effectively by leveraging the ability to manipulate, store, and retrieve the most relevant information based on an optimized indexing system. Essentially, this research explores how technology can be used to address real-life challenges that are affecting society today in a significant way, such as mining, recreating, and repurposing existing data to solve problems faced in automated smart city traffic monitoring and safety systems.
IoT-enabled infrastructure and interconnected devices are the most important elements that make up an integrated smart city and help address a variety of urban challenges by leveraging information and communication technologies. 
Over the past few decades, the field of autonomous technologies has witnessed significant progress, particularly in terms of speed and bandwidth enhancements. However, a notable discrepancy persists in the development of technology that precisely caters to user requirements, especially in specialized areas such as smart city transportation. Present approaches to road traffic accident detection demonstrate a shortfall in the sophistication and full operational effectiveness of intelligent transportation and accident detection systems. These datasets are designed to leverage the latest advancements in state-of-the-art algorithms.
Our research contributes to the evolving field of action recognition and computer vision, addressing a critical urban challenge: improving safety and efficiency within smart cities through the strategic implementation of these advanced technologies. By developing and publicly releasing a specialized \href{http://ieee-dataport.org/documents/comprehensive-dataset-ai-driven-traffic-accident-detection-and-computer-vision-systems}{dataset} focused on accident detection, our study marks a pivotal contribution to this domain. This initiative is expected to foster further academic research and also enhance practical applications in this essential sector, thereby contributing significantly to the technological evolution of smart urban environments.

\section{Overview of Existing Accident Detection Datasets}
\begin{table*}[h]
\small 
\centering
\caption{Overview of Traffic Accident Detection Datasets: Features, Access, and Collection Methods for Smart City Applications.}\label{tab:my-tablerq3}
\begin{tabular}
{|C{0.15\textwidth}|C{0.20\textwidth}|C{0.24\textwidth}|C{0.14\textwidth}|C{0.14\textwidth}|}
\hline
\textbf{Authors} & \textbf{Dataset} & \textbf{Dataset Feature} & \textbf{Access} & \textbf{Data Collection Approach} \\ \hline

Yao et al. \cite{Yao2022}         & DoTA (Detection of Traffic Anomaly)                       & 4,677 videos                                                                           &  Yes-\href{https://github.com/MoonBlvd/Detection-of-Traffic-Anomal}{link} & Dashcam                       \\ \hline
Yu et al. \cite{yu2021deep}       & Traffic Accident Data, Taxi GPS Data, Meteorological data & Varied weather (Cloudy, Snow, etc.), Road network and PoI                              & NA                & Sensors, Traffic Surveillance \\ \hline
Wang et al. \cite{Wang2020a}       & PEMS-BAY, META-LA                                         & 53116 videos from 325 sensors, 34272 videos from 207 sensors                           &  Yes-\href{https://pems.dot.ca.gov/}{link}   & Sensors, Traffic Surveillance \\ \hline
Bao et al. \cite{Bao2020}        & CCD (Car Crash Dataset), DAD and A3D                      & 6.35 hours, Varied weather conditions (snow, day and night), 2.43 hours and 3.56 hours &  Yes-\href{https://github.com/Cogito2012/UString}{link} & Dashcam                       \\ \hline
Reddy et al. \cite{Reddy2021}       & Traffic Driving data                                      & 182 drive sequences                                                                    & NA                & Dashcam                       \\ \hline
 Fernández et al. \cite{Fernandez-Llorca2020}  & PREVENTION dataset                                        & 6 hours video (80 meters around ego-vehicle), 3 radars 2 cameras and 1 LiDAR           & Yes-\href{https://prevention-dataset.uah.es/}{link}     & LiDAR Radar, DashCam          \\ \hline
Ali et al. \cite{Ali2022}       & TaxiBj, Bike NYC                                          & 16 months video recordings                                                             & NA                & Sensors                       \\ \hline
Wang et al. \cite{Wang2020}       & Spatial-Temporal Mixed Attention Graph-based Convolution model (STMAG)                & 2000 videos                                                                            & No-Future Release & Dashcam                       \\ \hline
Alkandari et al. \cite{Alkandari2015}  & Dynamic Webster with dynamic Cycle Time algorithm (DWDC)                                                       & \textbf{-}                                                                                      & NA                & Sensors                       \\ \hline
 Riaz et al. \cite{Riaz2022}       & KITTI, HTA, D2City                                        & 600 frames from Kitti, 286 clips, 65 frames and 678 video                              &  Yes-\href{https://github.com/MoonBlvd/Detection-of-Traffic-Anomal}{link}        & Dashcam                       \\ \hline
 Tang et al. \cite{Tang2017}     & 40-seconds traffic video                                  & 600*800 frame dimension                                                                & NA                & Traffic Surveillance          \\ \hline
 Huang et al. \cite{Huang2020}       & Traffic Management Centers report, IOWA DOT radar sensors & 856 crash reports, 29 sensors                                                          & Yes- \href{https://mesonet.agron.iastate.edu/request/rwis/traffic.phtml\\ }{link}       & Sensors                       \\ \hline
Bortnikov et al. \cite{Bortnikov2020}  & Video game GTA V, YouTube Car Accident Video              & 5 hours recording                                                                      & NA                & Traffic Surveillance, Dashcam \\ \hline
 Gupta et al. \cite{Gupta2021}     & DETRAC dataset                                            & 10 hours recording of 376 videos, 99 frames selected from each video                   &  Yes-\href{https://detrac-db.rit.albany.edu/download\\ }{link}        & Traffic Surveillance          \\ \hline
 Yang et al. \cite{Yang2021}       & Highway vehicle dataset, ImageNet VID dataset             & 32938 vehicle samples, 5354 videos                                                     & Yes-\href{https://image-net.org/download.php}{link}        & Traffic Surveillance          \\ \hline
 Ijjina et al. \cite{Ijjina2019}     & YouTube Accident Videos                                   & 20 seconds video chunks, Varied weather (harsh sunlight, daylight hours)               & NA                & Traffic Surveillance          \\ \hline
You et al. \cite{You2020}       & Causality in Traffic Accident (CTA)                       & 9.53 hours video from 1935 videos, 18 semantic cause and 7 semantic effect labels      &  Yes-\href{https://github.com/tackgeun/CausalityInTrafficAccident}{link}      & Dashcam                       \\ \hline
Srinivasan et al. \cite{Srinivasan2020}  & CADP                                                      & 1416 Accident footage                                                                  &  Yes-\href{https://ankitshah009.github.io/accident\_forecasting\_traffic\_camera}{link}       & Traffic Surveillance          \\ \hline
 Hui et al. \cite{Hui2015}        & -                                                         & -                                                                                      & NA                & Sensors                       \\ \hline
 Min et al. \cite{Xia2018}        & QMUL Junction dataset, AVSS dataset                       & 52 mins traffic video, 4 seconds chunks                                                &  Yes-\href{https://personal.ie.cuhk.edu.hk/$\sim$ccloy/downloads\_qmul\_junction.html\\ }{link}        & Traffic Surveillance          \\ \hline
 Vatti et al.\cite{Vatti2018}      & -                                                       &      -                                                                                 &             NA      &  sensors                       \\ \hline
\end{tabular}
\end{table*}

Our preliminary research focused on the review of action recognition systems in smart cities \cite{Adewopo2022ReviewOA}. The findings profiled datasets employed for accident detection in smart cities. As illustrated in Table \ref{tab:my-tablerq3}, these datasets encompass a range of features, types of sensors or video data, and access links for public usage.
\textit{Yao et al.}~\cite{Yao2022} constructed a benchmark dataset to evaluate traffic accident detection and anomaly detection across nine distinct action classes. This development highlights a critical gap in the field: the scarcity of extensively annotated, real-life accident datasets. To address this, \textit{Bortnikov}~\cite{Bortnikov2020} utilized simulated game video data, incorporating a variety of weather and scene conditions. This simulation-based approach yielded results comparable to real-life traffic videos found on platforms like YouTube, as detailed in Table~\ref{tab:my-tablerq3}.
Notably, most datasets in this domain, particularly for accident detection and autonomous vehicle research, are derived from sources like dashcams, traffic surveillance cameras, drones (such as the HighD, InD, or Interaction datasets~\cite{krajewski2018highd, zhan2019interaction}), and building-mounted cameras. The \textit{NGSIM HW101} and \textit{NGSIM I-80} datasets~\cite{colyar2007ngsim,halkias2006ngsim}, for instance, consist of 45-minute-long image sequences captured from eight synchronized cameras mounted on a building, recording at 10 Hz.
Despite their utility, \textit{Fernandez-Llorca et al.}~\cite{Fernandez-Llorca2020} argue that some datasets, such as NGSIM HW101, may not be entirely suitable for onboard detection applications. They are, however, invaluable for understanding and evaluating vehicle and driver behavior under varying traffic conditions. The \href{http://www.poss.pku.edu.cn/download}{PKU dataset} expands on this by offering over 5700 environmental trajectories collected using 2-D LiDAR with $\mathrm{360^\circ}$ coverage. It includes extensive vehicle trajectory data across 64 km and 19 hours of footage~\cite{zhao-pku}.
Another noteworthy dataset is the \href{https://prevention-dataset.uah.es/}{Prevention dataset}, encompassing data from three radars, two cameras, and a LiDAR unit. This dataset provides an 80-meter coverage radius around an ego-vehicle, supporting intelligent system development for vehicle detection and tracking \cite{izquierdo2019prevention}. The \href{http://apolloscape.auto/#}{Apolloscape dataset} is similarly significant for automatic driving and navigation in smart cities. It contains approximately 100K image frames and 1000km of trajectories, collected using four cameras and two laser scanners equipped with 3D perception LiDAR \cite{wang2019apolloscape}.
\textit{Ijjina et al.}~\cite{Ijjina2019} compiled a unique dataset comprising surveillance videos recorded at 30 frames per second, trimmed to 20-second clips. These videos were sourced from CCTV cameras at road intersections across various global locations, under diverse ambient conditions such as harsh sunlight, snow, and night hours.
Our analysis accentuates the diversity and significance of datasets in accident detection and autonomous vehicle research. These datasets, sourced from dashcams, surveillance cameras, drones, and building-mounted cameras, vary in features but collectively provide a comprehensive view of environmental trajectories, vehicle data, and diverse ambient conditions.
To enhance the applicability and performance of action recognition models in accident detection, future research should focus on utilizing a wide array of datasets. These should ideally encompass varied traffic and weather conditions, as well as diverse geographical locations, as demonstrated by \textit{Yu} and \textit{Bortnikov}~\cite{Bortnikov2020}. This approach not only improves model performance but also ensures real-world applicability.
Moreover, there is a pressing need for benchmark datasets that facilitate fair comparisons across different models and techniques in traffic accident detection. While simulated game video data has shown promise~\cite{Bortnikov2020}, the prioritization of real-life data is crucial for the practical effectiveness of developed models. The scarcity of extensively annotated real-life accident datasets presents a challenge for researchers.

\begin{table*}[h]
\footnotesize
	\caption{Distribution of traffic and dashcam footage across different data types.}
    \label{TrafficDataCount}
    \begin{tabular}{|
    >{\columncolor[HTML]{FFFFFF}}m{1.cm}|
    >{\columncolor[HTML]{FFFFFF}}m{1.4cm}|
    >{\columncolor[HTML]{FFFFFF}}m{1.5cm}|
    >{\columncolor[HTML]{FFFFFF}}m{1.5cm}|m{1.4cm}|m{1.2cm}|m{1.2cm}|m{1.5cm}|m{1.5cm}|m{1.2cm}|}
    \hline
   {\textbf{Data Type}} & {\textbf{Trafficam Accident}} & {\textbf{Trafficam Normal-Traffic}} & {\textbf{Total Trafficam Dataset}} & {\textbf{Dashcam Accident}} & {\textbf{Dashcam Normal-Traffic}} & {\textbf{Dashcam Total Dataset}} & {\textbf{Ext. Data Accident}} & {\textbf{Ext. Data Normal-Traffic}} & {\textbf{Total DataSet}} \\ \hline
    Train & 603 & 511 & 1114 & 763 & 639 & 1402 & 1250 & 146 & 3912 \\ \hline
    Val & 190 & 140 & 330 & 268 & 224 & 492 & 150 & 82 & 1054 \\ \hline
    Test & 134 & 60 & 194 & 196 & 154 & 350 & 100 & 81 & 725 \\ \hline
    \end{tabular}
\end{table*}

\section{Dataset Collection}
A comprehensive data collection process was implemented due to the lack of readily available robust datasets in this domain. This section provides a detailed description of our collection methodology and data sources.
To curate a robust, diverse dataset, we collected videos from a wide array of sources and different geographic locations worldwide. The majority of our video data originates from platforms like YouTube and Pexels. We employed strategic keyword searches to identify and curate relevant videos, including terms like "Car Crash", "Traffic accident," "Car accidents," and "Dashcam accidents." In order to maximize the geographic and linguistic diversity of our dataset, these keyword searches were conducted in multiple languages, including English, French, Spanish, and Russian.
Table [\ref{TrafficDataCount}] provides a detailed breakdown of traffic and dashcam videos, categorized by accident occurrences and normal traffic conditions. The data is further divided into training, validation, and testing sets. Table [\ref{TrafficDataCount}] showcases the distribution of these categories across different data types, including Trafficam, Dashcam, and External Data sources. In order to supplement our curated dataset, we utilized an external data source (Car Crash Dataset) publicly available on GitHub \cite{BaoMM2020}. We aggregated approximately 5,700 datasets to train and evaluate our model performance. The dataset created is publicly available at \href{http://ieee-dataport.org/documents/comprehensive-dataset-ai-driven-traffic-accident-detection-and-computer-vision-systems}{IEEE DataPort} \cite{adewopodataset}. 
\subsection{Traffic/Surveillance Data (Trafficam)}
This data contains videos captured from traffic and surveillance cameras situated at various locations. Trafficam provides a unique vantage point for capturing vehicular movement. These cameras are strategically positioned at various locations to offer an aerial or elevated view of the road, enabling a holistic view of vehicles. This perspective is essential for several reasons, including;
\textbf{Trajectory Angle}: Trafficam videos capture the trajectory angle of vehicles. This angle denotes the path and direction of a vehicle's movement, which is valuable for understanding vehicular dynamics and causal factors leading to an accident.
\textbf{Full Car View}: The elevated vantage point of traffic and surveillance cameras provides a full silhouette and profile of vehicles crucial for detecting anomalies or changes in vehicular posture, like tilting during a potential rollover.
Datasets from the TrafficCam angle are particularly beneficial for accident detection because the overhead view minimizes occlusions, allowing for an unobstructed view of potential accident sites, and monitoring multiple vehicles simultaneously that can help in detecting and analyzing multi-vehicle collisions.
\subsection{Dash Camera Video (DashCam)}
These videos are typically recorded from cameras installed on vehicle dashboards. DashCam videos offer a ground-level, front-facing perspective from vehicles, capturing the road ahead and occasionally vehicle interior/rear view. While invaluable in many respects, DashCams are beneficial for providing firsthand accounts of incidents and useful for understanding drivers' accident viewpoints, including capturing close-up details of incidents. Some of the challenges of Dashcams are the restricted field of view (focus mainly on the road ahead), Camera view occlusion, and variability in video quality based on different brands and models.

\section{Dataset Classes}

The dataset classes in our study are meticulously designed to encompass a wide range of traffic scenarios, focusing on various types of accidents as well as normal traffic conditions. This classification not only aids in a comprehensive understanding of traffic dynamics but also serves as a valuable resource for developing and testing accident detection and prediction models. Below, we detail each class and its significance in the dataset \cite{adewopo2023smart, adewopo2023baby}.

\subsection{Backend}
The `Backend' class represents rear-end collisions, where one vehicle strikes the back of another. This type of accident is common in traffic scenarios and is crucial for studying factors like sudden stops, tailgating, and distraction-induced accidents. Our dataset includes a significant number of these incidents, providing a robust basis for analyzing rear-end collision dynamics.

\subsection{Backend Rollover}
`Backend Rollover' extends the Backend class by including accidents where a rear-end collision leads to a rollover. This class is particularly important for understanding the severe impacts that can result from high-speed rear-end collisions and the dynamics of vehicle flips following such impacts.

\subsection{Frontend}
The `Frontend' class focuses on frontal collisions. These incidents are critical for studying head-on impacts and their consequences, including the effectiveness of safety features like airbags and crumple zones in vehicles.

\subsection{Frontend Rollover}
Similar to the Backend Rollover, the `Frontend Rollover' class includes frontal collisions resulting in a vehicle rollover. This class provides insights into the kinematics of rollovers following a frontal impact, which is vital for vehicle safety design and accident reconstruction studies.

\subsection{No Accident Normal Traffic}
The `No Accident Normal Traffic' class is essential for providing a baseline of normal driving conditions. This class includes data depicting typical, accident-free traffic scenarios, which is crucial for models to learn what constitutes non-accident conditions and improve their accuracy in accident detection.

\subsection{Sidehit}
The `Sidehit' class encompasses accidents involving a side-impact collision. These are particularly important for studying the effects of lateral impacts on vehicle occupants and the structural integrity of vehicles.

\subsection{Sidehit Rollover}
Expanding on the Sidehit class, the `Sidehit Rollover' includes side-impact collisions that lead to rollovers. This class is critical for understanding the dynamics and risks associated with rollover accidents following a side impact.

\subsection{General Augmented Crash}
The `General Augmented Crash' class is a composite category that includes various types of accidents not specifically covered in the other classes collected from external data source (Car Crash Dataset) publicly available on GitHub \cite{BaoMM2020}. This class is particularly useful for studying less common but equally important accident scenarios and for generalizing the model's ability to recognize diverse accident types.

\subsection{Data Distribution}
Our dataset is divided into training, validation, and testing sets to facilitate effective model training and evaluation. The training set, with 3912 instances, includes 2616 accident scenes, ensuring that models learn to distinguish between normal and abnormal traffic conditions effectively. The validation set comprises 1054 instances, with 446 depicting normal traffic, aiding in fine-tuning the models. The testing set, consisting of 725 instances with 420 representing traffic-accident scenarios, is crucial for evaluating the model's performance in real-world conditions.

\subsection{Annotation and Usage}
Each video segment in our dataset has been meticulously annotated using the Labelbox tool, providing detailed information about the accident type. This rich annotation framework enables researchers and developers to train sophisticated machine-learning models for various applications, including accident detection, autonomous vehicle navigation, and traffic management systems. The diverse range of classes and the balanced representation of accidents and normal traffic scenarios make this dataset a comprehensive resource for advancing research in traffic safety and vehicle automation.
To ensure our dataset is concise and relevant, videos were processed based on the rapid dynamics of accidents. We segmented the videos into five second non-overlapping clips, ensuring that each segment captures a distinct event or scene. This segmentation strategy aids in minimizing redundancy and focusing on the most relevant content for accident recognition.
Each five-seconds video segment was annotated manually using the Labelbox annotation tool.

\section{Conclusion and Discussion}
In this paper, we explored enhancing traffic accident detection within the burgeoning landscape of smart cities. Through the collection and classification of data from Traffic/Surveillance cameras and Dash camera videos, we curated a robust dataset that diversifies the range of accident scenarios and also enriches the field of accident detection research. This dataset contains a variety of classes annotated with a level of precision that paves the way for advanced research and practical applications in this domain.

The cornerstone of our research is the implementation of the I3D-CONVLSTM2D model, which utilizes the dataset curated in this study \cite{Adewopo2022ReviewOA,adewopo2023ai}. Integrating RGB frames with Optical Flow in the research presented in \cite{adewopo2023smart,adewopo2023ai}, demonstrates the potential of our dataset in enhancing the accuracy and efficiency of accident detection systems. The development of such models is crucial in the context of smart cities, where timely and accurate accident detection can significantly impact public safety and traffic management.

Furthermore, the public availability of our benchmark dataset stands as a testament to our commitment to advancing the field. By making this dataset accessible, we aim to foster a collaborative environment where researchers and practitioners can build upon our work, leading to continuous improvements and innovations in accident detection technologies. This open-access approach accelerates research in the field and also ensures a wider range of perspectives and methodologies are applied, enriching the overall quality and applicability of the research.

The integration of these advanced systems into existing urban infrastructure is another critical aspect of our work. This integration is essential for maximizing the impact of our research on public safety and for paving the way toward a more responsive and efficient urban environment. In our data curation, we address ethical and privacy concerns associated with the deployment of these technologies. Ensuring that these systems align with societal values and norms is crucial for their acceptance and effectiveness.

In conclusion, our research represents a significant stride towards realizing smarter, safer cities. The path forward is ripe with opportunities for innovation, necessitating a collaborative effort among researchers, city planners, and policymakers. Harnessing the power of AI and machine learning, as demonstrated in our study, brings us closer to transforming urban environments into safer, more efficient spaces. The potential for groundbreaking advancements inherent in this research is vast, and its implications for the future of urban living are profound. As we continue to explore and innovate, our work lays a solid foundation for the next generation of traffic accident detection systems, ultimately contributing to the creation of smarter and safer urban communities.
\bibliography{Main}
\bibliographystyle{ieeetr}

\end{document}